\documentclass[10pt,twocolumn,letterpaper]{article}

\usepackage{cvpr}
\usepackage{times}
\usepackage{epsfig}
\usepackage{graphicx}
\usepackage{amsmath}
\usepackage{amssymb}

\usepackage[pagebackref=true,breaklinks=true,letterpaper=true,colorlinks,bookmarks=false]{hyperref}

\cvprfinalcopy 

\def\cvprPaperID{19} 

\ifcvprfinal\fi
\begin{document}

\title{Source Generator Attribution via Inversion}

\author{Michael Albright and Scott McCloskey\\
Honeywell ACST\\
1985 Douglas Drive North, Golden Valley, MN, USA\\
{\tt\small firstname.lastname@honeywell.com}
}

\maketitle

\begin{abstract}
With advances in Generative Adversarial Networks (GANs) leading to dramatically-improved synthetic images and video, there is an increased need for algorithms which extend traditional forensics to this new category of imagery.  While GANs have been shown to be helpful in a number of computer vision applications, there are other problematic uses such as `deep fakes' which necessitate such forensics.  Source {\em camera} attribution algorithms using various cues have addressed this need for imagery captured by a camera, but there are fewer options for synthetic imagery.  We address the problem of attributing a synthetic image to a specific generator in a white box setting, by inverting the process of generation.  This enables us to simultaneously determine whether the generator produced the image {\em and} recover an input which produces a close match to the synthetic image.  
\end{abstract}

\section{Introduction}

Because of its use in `fake news' and `revenge porn', the implications of fully- or partially-synthetic imagery has recently become a matter of broad social concern.  Underlying both of these is the technology of deep networks used to generate imagery, often of faces, that are increasingly realistic.  Whereas traditional image forensics already include powerful techniques applicable to images captured with a wide range of cameras \cite{Lukas,Kurusowa,ChenCamcorder,Choi,Kharrazi}, there are relatively fewer options available to forensic analysts operating on synthetic imagery.

We aim to close the capability gap between source {\em camera} attribution and source {\em generator} attribution, and to provide additional functionality for generator attribution.  Source camera attribution methods vary considerably, but typically use low-level cues (such sensor non-uniformity) and statistical indicators; they are unable to re-create the image capture process, because the physical camera may be unattainable and the scene being photographed may be ephemeral.  Neither of these restrictions apply to synthetically-generated imagery, so we expand the attribution problem to encompass both the determination of whether the generator produced the image {\em and} the inputs necessary to re-create the generation process.  Mathematically, we describe the generator as a function $G$ which transforms a vector $z$ into an image $I_g$ as 
\begin{equation}
I_g = G(z).
\end{equation}  
When there are multiple (known) generators $G_1$, $G_2$, etc. the {\bf limited attribution problem} is to determine for a `probe' image $I$ the value of $i$ that - for some value $z$ - satisfies $G_i(z) = \tilde{I} \approx I$.  The related {\bf inversion problem}, given a specific generator $G$ and probe image $I$, is to estimate a latent vector $\tilde{z}$ such that $G(\tilde{z}) = \tilde{I} \approx I.$  In both, we use approximation to acknowledge that small differences should be expected due to quantization, dynamic range clipping, and perhaps compression applied to $I$.  

\begin{figure}[t]
\begin{center}
\includegraphics[width=0.9\linewidth]{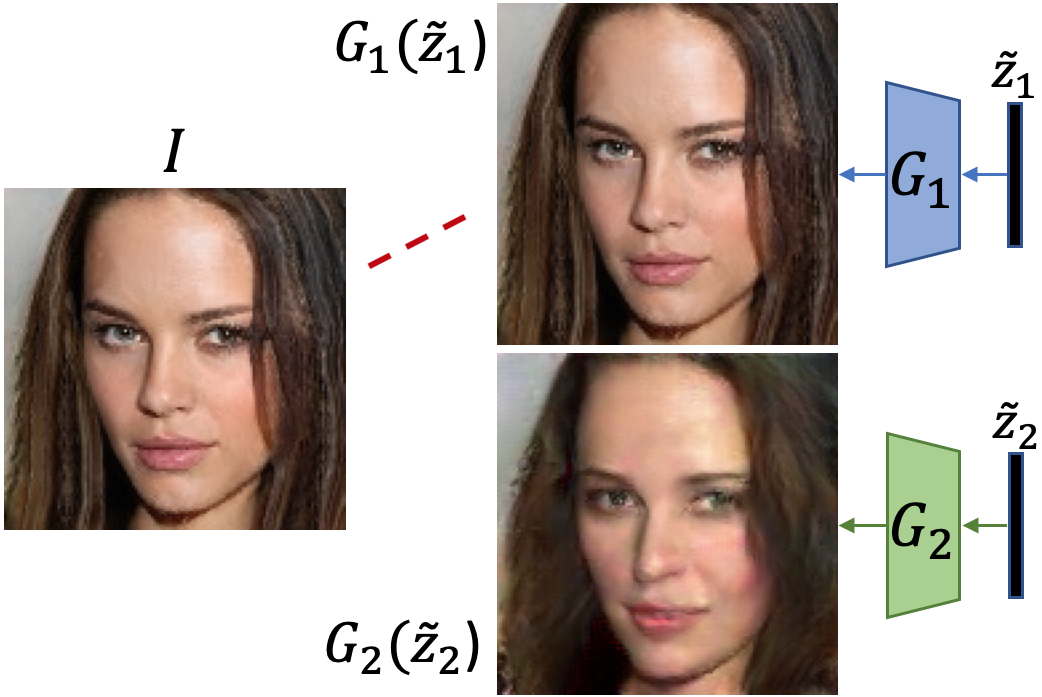} \\
\end{center}
\caption{Image attribution by generator inversion:  Given a synthetic probe image $I$ and pre-trained generators $G_i$, the generators are inverted to find latent vectors $\tilde{z}_i $ such that generated outputs $G_i(\tilde{z})$ approximate the probe.  Attribution is assigned to the generator with the best reconstruction (red dashed line).  }
\label{fig:image_attribution_diagram}
\end{figure}

In our {\bf extended attribution problem} (Fig. \ref{fig:image_attribution_diagram}), we consider the case where there are multiple (known) generators and, given a probe image, we determine {\em both} $i$ and $\tilde{z}$ such that $G_i(\tilde{z}) = \tilde{I} \approx I$.  The utility of attribution, in the context of synthetically generated imagery, is that training GAN-based generators for high-resolution outputs is compute-intensive, data-intensive, and error-prone.  As such, the majority of users are more likely to download and use publicly-available generators than they are to train their own from the ground up.  We also believe that, beyond limited attribution, the ability to re-create the generation process (enabled by our estimation of $z$) is critical to improving the explainability of the attribution decision.

In our experiments, we demonstrate extended attribution on generators from two different domains, trained via two different methods.  We first show that the generative layers of an auto-encoder, trained on MNIST \cite{Lecun98gradient-basedlearning} characters, can successfully be attributed in increasingly difficult scenarios, up to and including discrimination between generators trained on the same data, in the same order, but with different random seeds.  We then show that generators trained to produce realistic facial images in a GAN can be attributed, and discuss some of the interesting similarities and differences between the probe and reconstructed image.

\section{Related Work}

Both the limited attribution and inversion problems described in the introduction have attracted attention in recent years.  For attribution, Yu \etal \cite{yu_attributing_2018} and Marra \etal \cite{marra_attributing_2018} have recently presented methods which use noise-type signatures to attribute a given image to a specific generator, which is conceptually similar to how Photo Response Non-Uniformity (PRNU) \cite{Lukas} provides source {\em camera} identification.  These methods are very successful, providing nearly perfect attribution performance, even being able to discriminate between two generators with the same architecture, having been trained on the same data.  A related problem is the detection of whether a given image was generated by {\em any} GAN-type generator, which has successfully been demonstrated by several groups using various methods \cite{marra_detecting_2018,nataraj_detecting_2019,hsu,guera,albanyEyes,self_detecting_2019}.

The generator inversion problem has likewise attracted attention.  Creswell and Bharath \cite{creswell_inverting_2018} develop a generator inversion method as a means to explore GAN performance as a function of attributes in the image space.  Luo \etal \cite{luo_learning_2017} solve inversion by training an encoder coupled to a pre-trained generator in an auto-encoder framework.  Lipton and Tripathi \cite{lipton_precise_2017} introduce a method called stochastic clipping to recover arbitrarily precise approximations of $z$, even in the presence of simple types of noise.  None of the prior inversion work has addressed attribution.

Another related work, by Kilcher\etal \cite{kilcher_generator_2017}, demonstrates that generators instantiated with 
random weights can produce blurry versions of real images from various benchmark datasets.  This demonstrates that the convergence of inversion, by itself, does not signal that the image can be attributed to a generator.  

\section{Method}

Our method for attributing a probe image $I$ to a generator $G_i$ is predicated on the assumption that other generators $G_{j }$ $(j \ne i)$ cannot generate $I$ as well as the true generator $G_i$, so the minimum reconstruction error will correspond to the true generator $i$. We offer a brief discussion on why one may reasonably expect that situation to often (though not always) hold:  First, we note that generators map a low dimensional latent vector $z$ into very high dimensional output space $G_j(z)$.  For instance, in the case of ProGAN \cite{karras2017progressive}, dim($z$) = 512 and dim($I$) = 3,145,728  (3x1024x1024).  Hence, the mapping from $z$ to $G_j(z)$ parameterizes a union of low dimensional manifolds (of dimension $\le$ dim($z)$) in the high dimensional space \cite{arjovsky_towards_2017}.  Furthermore, there are a number of sources of inherent randomness in generator training, such as random initialization of network weights, ordering of training images, and absence or inclusion of particular images in training batches.  Arjovksy and Bottou \cite{arjovsky_towards_2017} have shown that it is highly improbable for low-dimensional manifolds in high-dimensional spaces to perfectly overlap everywhere if subjected to random perturbations; this is in fact part of the motivation for using a Wasserstein metric in Wasserstein GANs \cite{arjovsky_towards_2017,arjovsky_wasserstein_2017}.  Furthermore, in the case of GANs, training failures such as \textit{mode dropping} may make the generator incapable of generating some regions of the manifold of natural images.

In the extended attribution problem, illustrated in Figure \ref{fig:image_attribution_diagram}, we start from a probe image $I$ which was generated by one of several generators $G_1$, $G_2$, \dots, $G_n$ , where we assume we know both the architecture and weights of all generators.  We do not know a-priori which generator $G_i$ made the probe image, nor do we know the latent vector z, but we seek to determine them.  Note that if we could perfectly identify $z$ and $i$, we could (in the ideal case) perfectly recreate the probe image $I = G_i(z)$.  In practice, there may be some residual discrepancies due to post-processing of the generated image, e.g. quantization, clipping, image compression, etc., so we allow for for some small differences between the probe image and the generator output.  Hence, we seek to estimate a latent vector $\tilde{z}$ such that $G_i(\tilde{z}) = \tilde{I}$ is as close as possible to the true probe $I$.  For our experiments, we formalize generator inversion as an optimization problem, where we minimize the loss function
\begin{equation}
  L_j(z) = \frac{1}{MN} || I - G_j(z) ||^2  ,
  \label{eq:loss_fcn}
\end{equation}
with $N$ and $M$ being the number of pixels and color channels in the image, respectively.  

We separately minimize Eq. \ref{eq:loss_fcn} for each generator $j$, using an optimization algorithm to find the best latent vector $z$ which minimizes the loss function.  Attribution is assigned to the generator $i$ with the smallest residual error, and the estimated latent vector $\tilde{z}$  is the point of minimum loss
\begin{equation}
  \tilde{z} = \operatorname*{argmin}_z \frac{1}{MN}  || I - G_i(z) ||^2  \,.
\end{equation}

Because generators used in GANs and autoencoders are neural networks, they are readily differentiable via backpropagation, so the loss function in Eq. \ref{eq:loss_fcn} may be optimized efficiently using gradient-based methods; in this work, we found the Adam \cite{kingma2014adam} optimization algorithm to perform well for all tested generators.
Because Eq. \ref{eq:loss_fcn} is not convex, there is no guarantee that the optimization will converge to a global minimum.  Hence, to ensure low reconstruction errors are obtained, we perform a multi-start optimization, where we perform each optimization multiple times, each starting from different random initial starting guesses for the latent vector $z$, and we choose the result with the lowest residual error.

We note that since the magnitude of the minimum residual error  $L_i^{\rm min} = L_i(\tilde{z} $)  quantifies how well the generated output $G_i(\tilde{z}) = \tilde{I}$ matches the probe image $I$, one may use the residual errors to assess confidence in the attribution assignment: for a correct assignment, one would expect the minimum loss $L^{\rm min}_i$ to be very small, and for high confidence it should be significantly smaller than reconstruction errors  $L^{\rm min}_j$ from other generators $j \ne i$. 

In the special case where there are only two generators $i$ and $j$, we can summarize our attribution decision and confidence by a single numerical score:
\begin{equation}
 S = \frac{ L^{\rm min}_j - L^{\rm min}_i }{L^{\rm min}_j + L^{\rm min}_i}  \, .
 \label{eq:Sequation}
\end{equation}
Note that $S \rightarrow 1$ when the $i^{\rm th}$ generator perfectly reconstructs the probe ($L_i^{\rm min} \rightarrow 0$),  $S \rightarrow -1$ when the $j^{\rm th}$ generator perfectly reconstructs the probe ($L_j^{\rm min} \rightarrow 0$), and $S \rightarrow 0 $ when both generators reconstruct the probe equally well ($L_i^{\rm min} = L_j^{\rm min}$; no attribution possible).  Hence, the score $S$ provides a natural way to evaluate attribution performance using Receiver Operating Characteristic (ROC) curves, effectively treating binary attribution as binary classification; we do so in Section \ref{subsec:MNIST_experiments}.

In the next section, we show that differences in generator training indeed contribute to detectable differences in generated images, even in the case of identical generator architectures.

\section{Experiments}

In this section, we describe experiments by which we demonstrate the utility of our generator attribution and inversion.  They present a series of increasingly difficult attribution problems, distinguishing between pairs of generators with increasing similarity in how they are trained.  

\subsection{MNIST Experiments}
\label{subsec:MNIST_experiments}

The first set of experiments were carried out with the MNIST dataset, selected because the smaller image size allows for faster training and testing.  In each MNIST experiment, we trained a pair of fully-connected auto-encoders with sigmoid activations and an L2 loss.  The encoder and decoder parts are symmetric, each having 784 nodes at the input/output layers and hidden layers with 64 and 32 nodes each.  Each auto-encoder was trained on approximately 30,000 MNIST digits for 20,000 steps in batches of 256 images using the Adam optimization algorithm
with a learning rate of 0.01.

Once the training was completed, the decoders were separated from their encoders and the weights were frozen, rendering them pre-trained generators which map 32 dimensional latent vectors into 784 dimensional outputs.  We then performed optimization-based generator inversion for image attribution on a test set consisting of 500 digits generated from each generator and saved as a PNG.  We again used the Adam optimization algorithm with a learning rate of 0.01, and ran it for 1000 steps per inversion.  Inversion of the MNIST generators proved straightforward, but to reduce the chance of optimizers getting trapped in local minima and biasing the results, we implemented a multi-start optimization strategy, where optimization was repeated 10 times per image with 10 different random initial starting guesses for the latent vector $z$.    We assess performance by treating attribution as a binary classification problem and plotting the Receiver Operating Characteristic (ROC) curve; we use the Area Under the Curve (AUC) as a performance summary statistic.  We arbitrarily designated one of the generators as the target generator and measure the True Positive Rate (TPR) at which outputs from the target generator are classified as such.  The False Positive Rate (FPR) measures the frequency with which non-target generator outputs are classified as target outputs.  These values are computed over a range of thresholds on $S$ (as defined in Eq. \ref{eq:Sequation}).  

Finally, we recall that each trained generator is influenced by a number of sources of randomness in the training process, such as the initial values of network weights, the order of images in training, etc., so attribution performance may vary if the experiment is repeated with different random number generator seeds.  To assess the uncertainty in attribution performance, we repeat each MNIST experiment five times and plot five ROC curves per figure.

\subsubsection{Non-overlapping Training Data}

\begin{figure}[t]
\begin{center}
\includegraphics[width=0.9\linewidth]{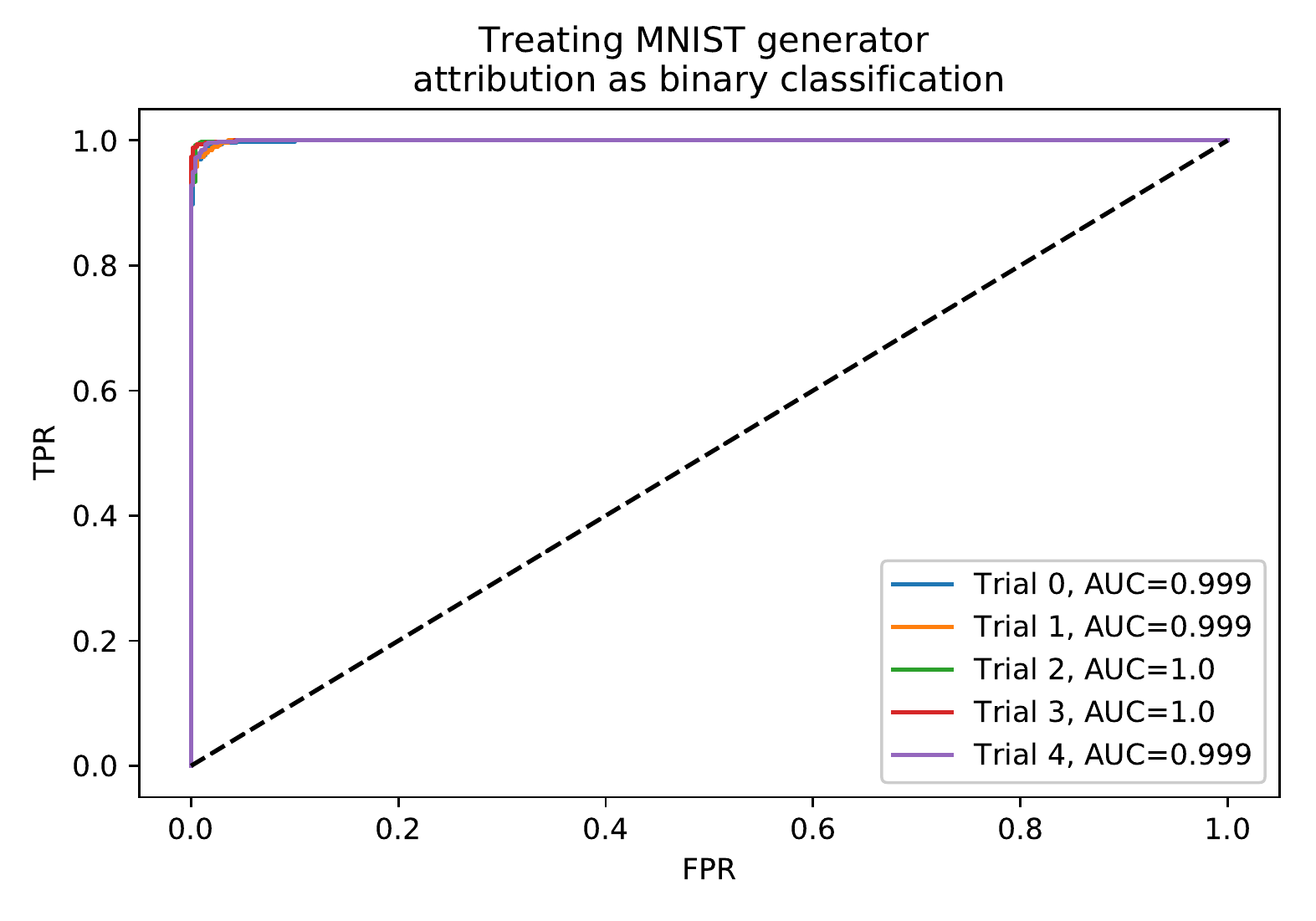} \\
\includegraphics[width=0.9\linewidth]{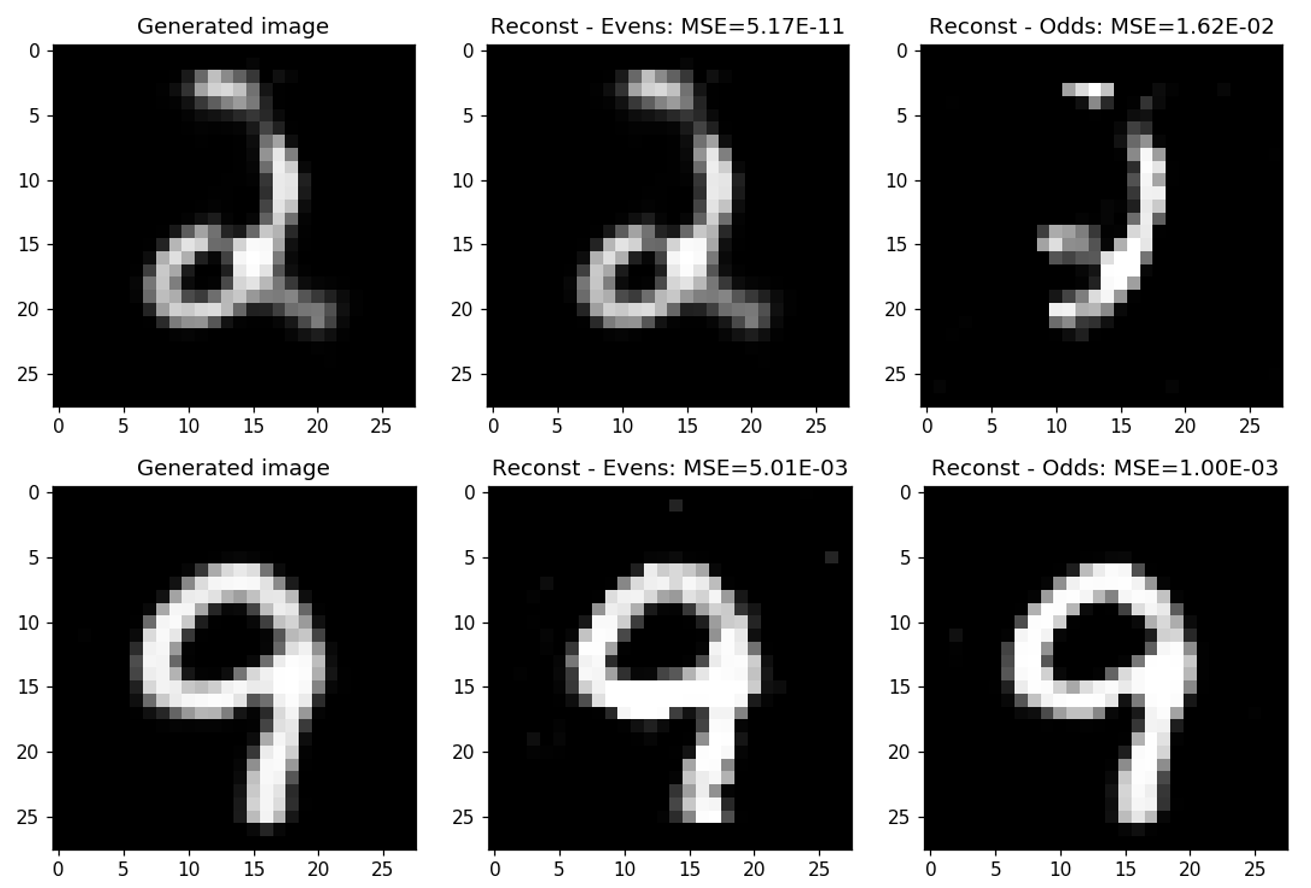} 
\end{center}
\caption{(Top) ROC curve on attribution between two generators with non-overlapping training data.  One generator is trained on only even MNIST digits, and the other on odd digits.  For reference, the diagonal line shows the performance of random attribution.  The experiment is repeated 5 times.  (Bottom rows) Examples of two successful attributions.  The probe image (left) was approximated with the even (middle) and odd (right) generators via inversion.}
\label{fig:ROC_oddEven}
\end{figure}

In this experiment, we trained the auto-encoders on two non-overlapping subsets of the MNIST digits.  For simplicity, we train one using odd digits and the other with even digits, though we note that our attribution does {\em not} recognize the digit or its parity.  Having thus trained even and odd digit generators, we attribute each of the test images by the method descried above.  Figure \ref{fig:ROC_oddEven} shows that the performance of our attribution on this experiment is nearly perfect, despite the fact that the generators do a surprisingly good job synthesizing digits that they've never seen before.  The bottom row of Figure \ref{fig:ROC_oddEven}, for example, shows a `9' synthesized by an odd generator (left column) and reconstructions from inverting both even and odd generators (center and right columns, respectively).  Despite never having been trained on `9's, the even number generator can produce a good approximation of this input, though less so for the odd generator and the `2' in the row above.  We also observe that variance in the area under the curve is quite low over the five different repetitions of the experiment, indicating the attribution performance is robust.

\subsubsection{Same Training Data, Different Order}
\label{sec:samedatadifforder}

\begin{figure}[t]
\begin{center}
\includegraphics[width=0.9\linewidth]{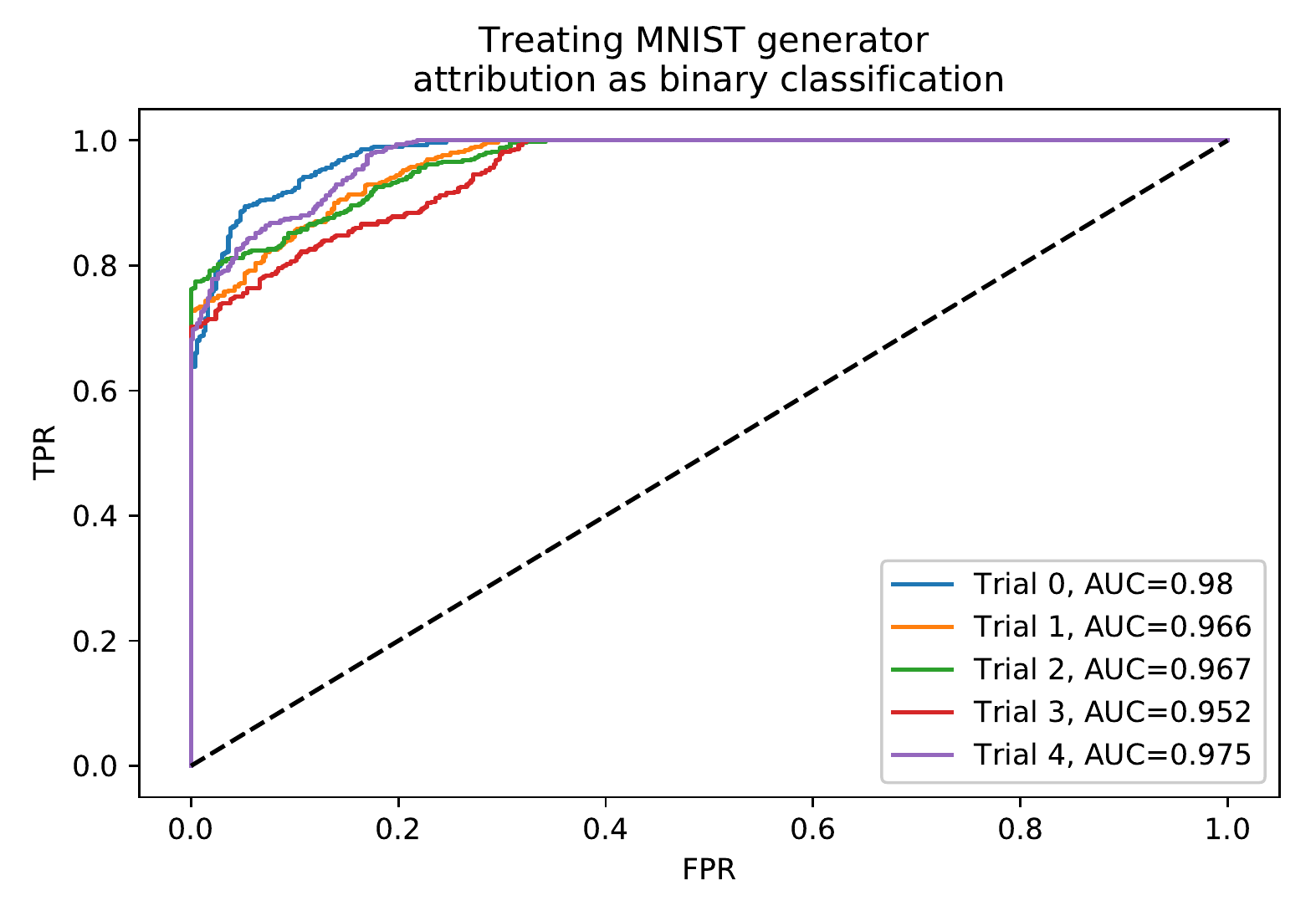} \\
\includegraphics[width=0.9\linewidth]{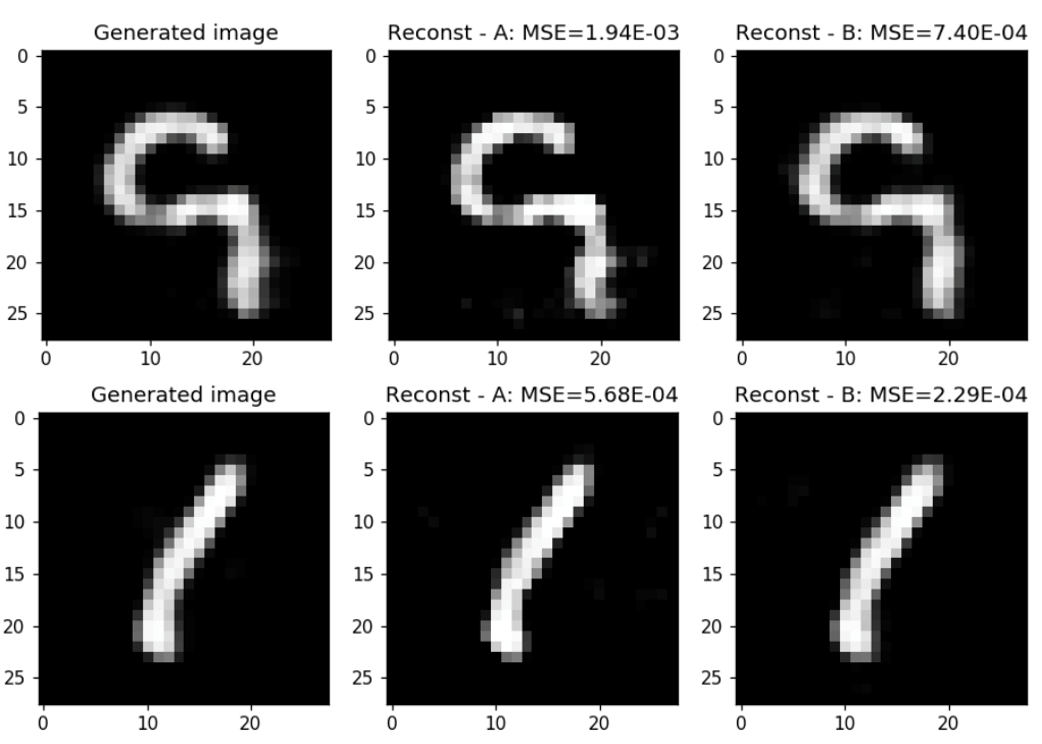} 
\end{center}
\caption{(Top) ROC curve on attribution between two generators with different permutations of the same training data.  The experiment is repeated 5 times. (Bottom rows) Examples of two {\em unsuccessful} attributions. }
\label{fig:ROC_shuffle}
\end{figure}

In this experiment, we trained the auto-encoders on the same set of MNIST digits (comprised of both even and odd digits), but shuffled the order in which they're used.  Both generators have the same architecture, but their weights receive different random initializations at the start of training.   Figure \ref{fig:ROC_shuffle} shows that the performance is reduced when our algorithm is asked to differentiate between generators trained with the same training data, and the variation in performance between trials has also increased, but the performance is still quite good in every trial.  The figure also shows two examples of {\em mis}-attribution, illustrating that the outputs of the two generators are quite similar, visually, and have reconstruction errors which are quite close to one another.  This portends difficulty in successfully attributing images subject to the normal forensic challenges such as compression and re-encoding.

\subsubsection{Same Data, Same Ordering}

\begin{figure}[t]
\begin{center}
\includegraphics[width=0.99\linewidth]{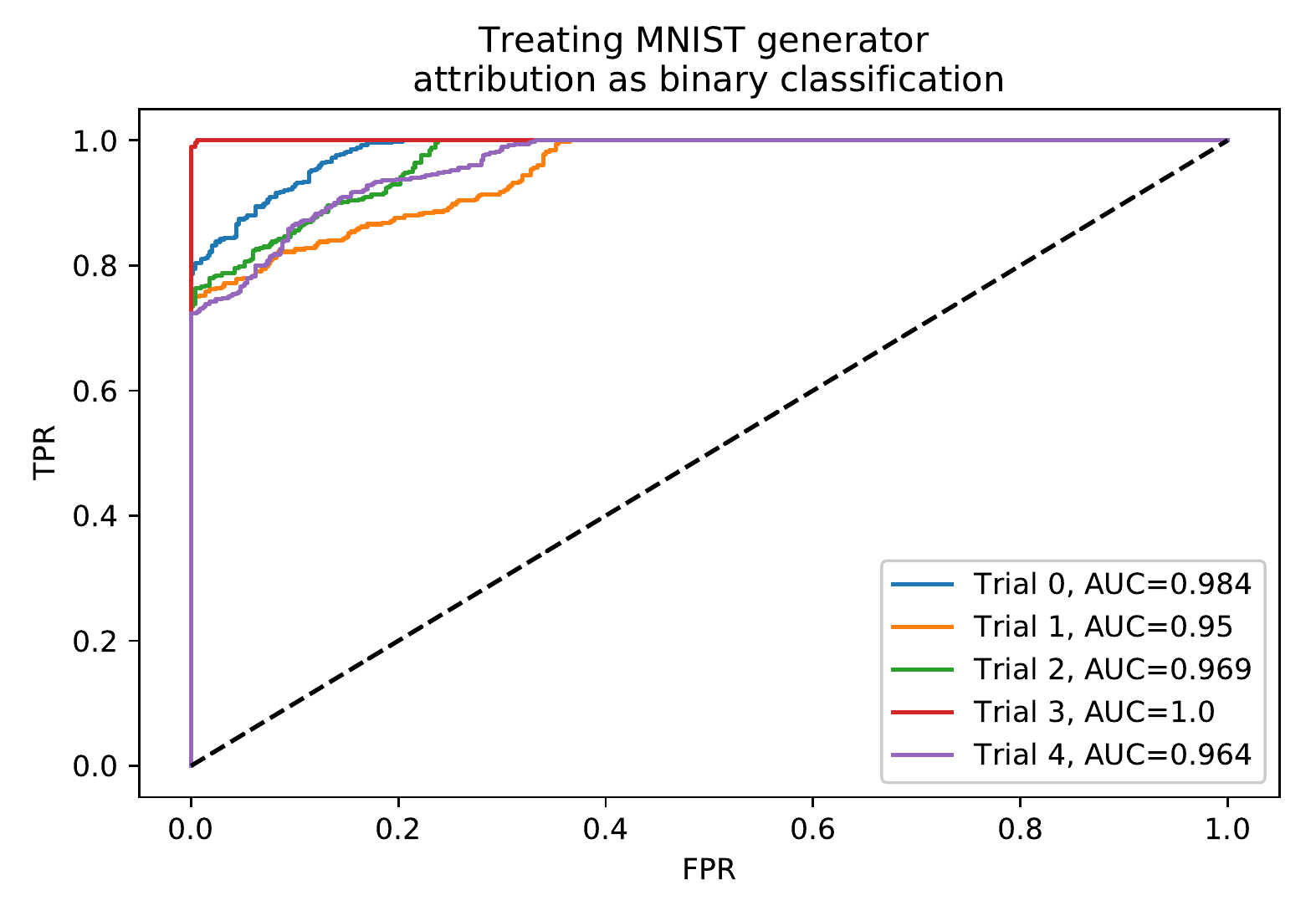} 
\end{center}
\caption{ROC curve on attribution between two generators with the same training data, in the same order. Only the random initialization of the network differs between the two generators, despite which we can attribute their outputs well.  The experiment is repeated 5 times.}
\label{fig:ROC_initialization}
\end{figure}

In this experiment, the two auto-encoders are trained on the same subset of MNIST digits, and those training digits are presented in the same order.  The only difference between the two generators is the initial (random) weights.  Despite this high level of similarity, Fig. \ref{fig:ROC_initialization} shows that our approach can unambiguously attribute $> 70\%$ of the inputs with extremely high confidence for all five trials. 

\subsubsection{Same Training Data, Different Order, with Compression}
\label{sec:compression}

In the last of our MNIST experiments, we quantify the impact of JPEG compression on the performance of the attribution algorithm.
Compression is a well-known nuisance factor for many forensic algorithms.  In the case of our attribution algorithm, compression introduces perturbations in the probe image which move it off of the manifold representing the span of the generator's output.  Note that even lossless compression does this, as the double precision output of the generator is quantized to an integer.  Given that (a) compression is expected in most real-world uses of attribution, (b) Kilcher \etal \cite{kilcher_generator_2017} showed that random GANs can approximate a probe image, and (c) our empirical observation that outputs from the wrong generator can be quite close to the probe image, it is important to understand how likely compression is to lead to mis-attribution.

\begin{figure}[t]
\begin{center}
\includegraphics[width=0.99\linewidth]{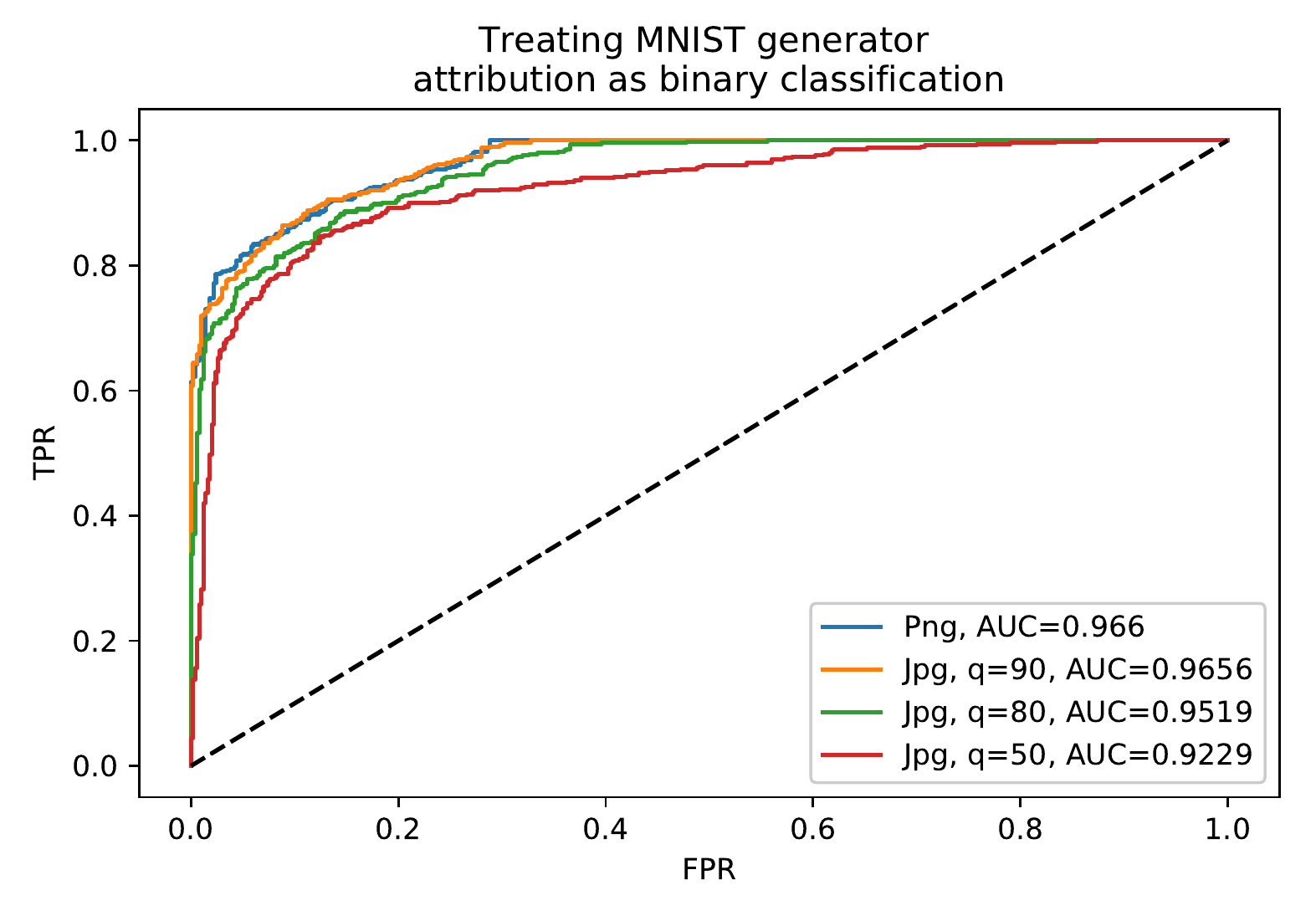}
\end{center} 
\caption{ROCs demonstrating performance of our attribution as a function of JPEG compression quality factor.  The blue curve represents performance on probe images saved as PNGs, and the other curves represent performance on JPEGs with a range of quality factors $q$.  While increasing compression reduces performance, the reduction is graceful and performance is still significantly better than chance at a quality factor of 50.}
\label{fig:ROC_compression}
\end{figure}

In order to study this impact, 
we re-visit the experiment described in Sec. \ref{sec:samedatadifforder}, where we discriminate between 
two generators trained on the same MNIST data, but with different orderings of the training data.   
After training a pair of networks, the two generators are used to create digits
which we compress using OpenCV's imwrite function, using different values for the quality factor $q$ (which ranges from 0 to 100).
Fig. \ref{fig:ROC_compression} shows ROC curves from this experiment, along with a reference ROC for images saved as PNGs without compression.  We see that while increasing levels of JPEG compression do reduce performance, the degradation is graceful and attribution performance is still decidedly better than chance.

We emphasize that the different ROC curves shown in Fig. \ref{fig:ROC_compression} all originate from a \textit{single} pair of generators,
and the differences between the curves originate in different compression levels applied when saving a common set of test images.  This is very different from Fig. \ref{fig:ROC_shuffle}, which tested attribution performance variation across \textit{different} pairs of generators trained with different random initializations and different orderings of training data (but without lossy compression).

\begin{figure}[t]
\begin{center}
\includegraphics[width=1.01\linewidth]{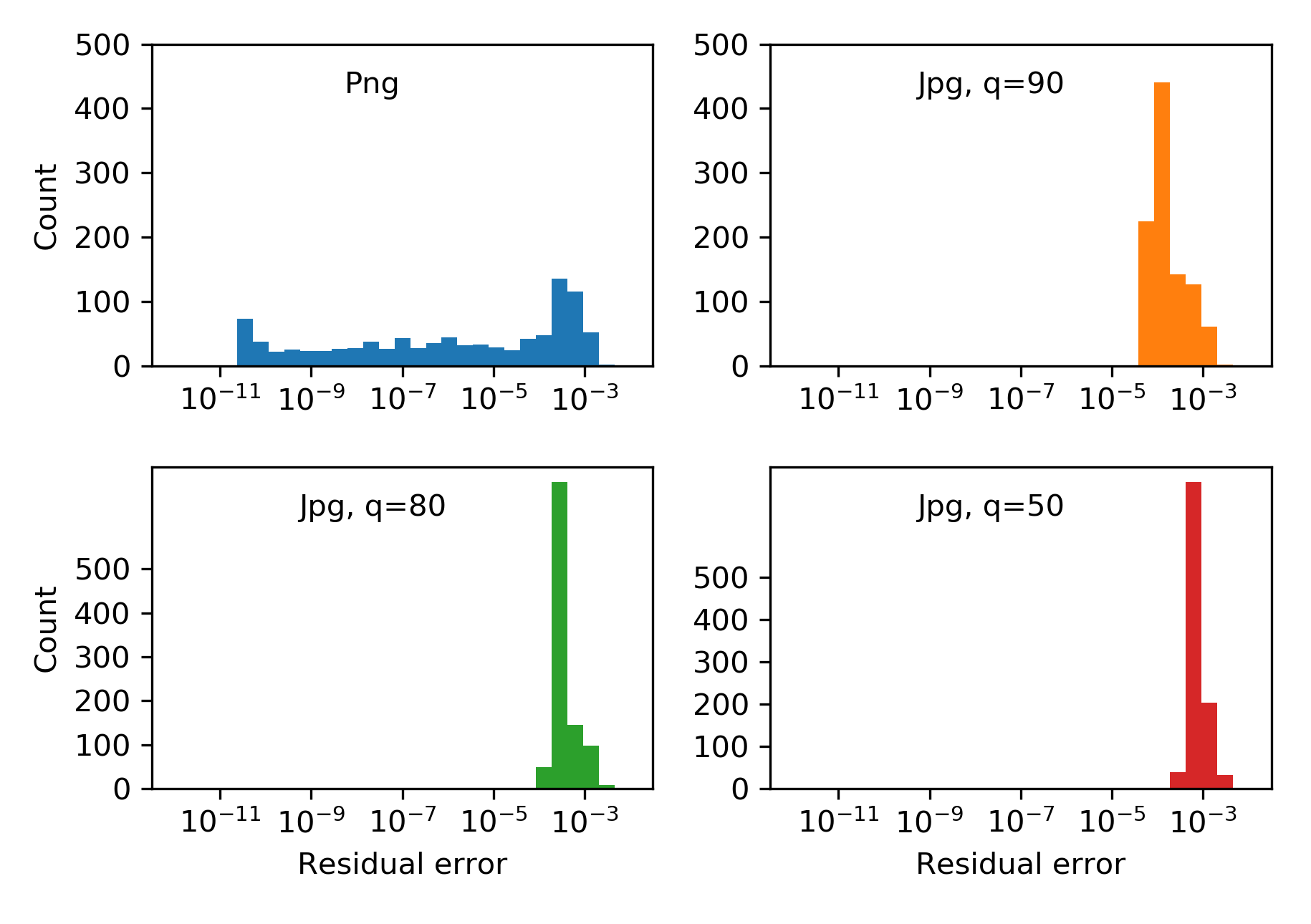}
\end{center}
\caption{Histograms of reconstruction errors (Eq. \ref{eq:loss_fcn}) at different compression levels.  The plots show reconstruction errors obtained when inverting each of the  generators from Sec. \ref{sec:compression} on the test images created by that generator. }
\label{fig:CompressionErrorsAB}
\end{figure}

In Fig. \ref{fig:CompressionErrorsAB}, we plot histograms of reconstruction errors (Eq. \ref{eq:loss_fcn}) at different compression levels.  In each histogram, we show the errors obtained when inverting each of the  generators on the  test images produced by that generator.  We see reconstruction errors increase as the quality factor $q$ decreases.  However, the errors are still low enough to permit attribution, as evidenced by the large AUCs in Fig. \ref{fig:ROC_compression}.

\subsection{CelebA Experiments}
\label{subsec:CelebA_experiments}

In the second set of experiments, we performed attribution on generators that were trained as Generative Adversarial Networks (GANs).
The three GANs tested were ProGAN \cite{karras2017progressive}, SAGAN \cite{zhang2018SAGAN}, and SNGAN \cite{miyato2018spectral}, all of which were modified from their original description to output 128x128 resolution images and were trained on face images from the CelebA dataset.  We re-used pre-trained generator weights shared by Ning Yu \cite{yu_attributing_2018}.  The SAGAN and SNGAN generators proved easy to invert via optimization algorithms, so multi-start was not strictly necessary, but for good measure we employed 10 random initial optimization starts per image. For each random start, we used the Adam optimization algorithm with a learning rate of 0.1 and 1000 optimization steps per image, and chose the best reconstruction with the lowest loss encountered.  When inverting SAGAN, we found it helpful to employ an explicit learning rate reduction on plateau, to reduce oscillation in the loss as the optimizer approached the minimum; we used a learning rate shrink factor of 0.5 and a patience of 30.  The ProGAN generator proved slightly more challenging to invert reliably, so we employed 20 random initial optimization starts per image. We also used the Adam optimizer, but with a learning rate of 0.9, 300 optimization steps per image, and no explicit learning rate reduction (besides those built into Adam).

\begin{figure}[t]
\begin{center}
\includegraphics[width=0.9\linewidth]{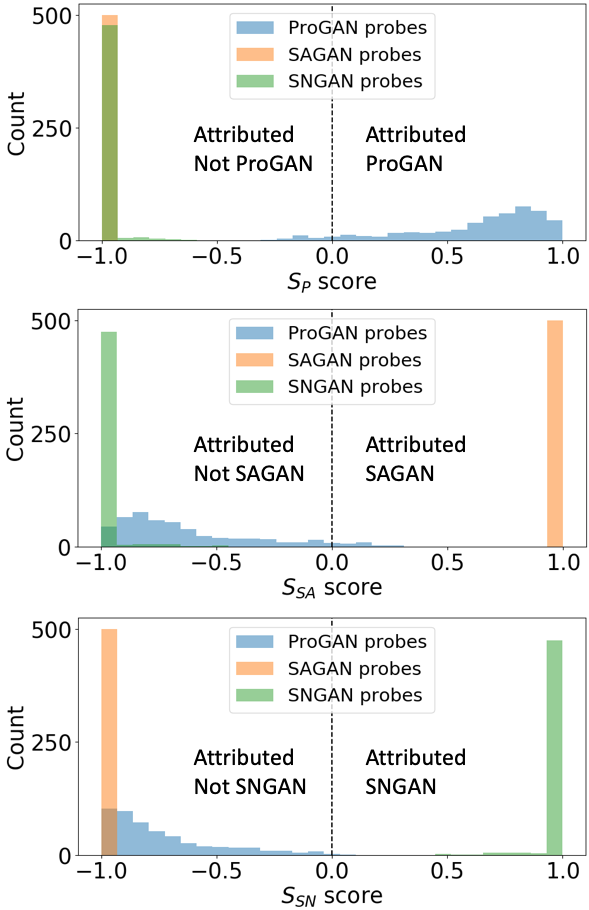} 
\end{center}
\caption{Histograms of $S_{i}$ scores for each generator.  The top figure evaluates inversion of the ProGAN generator $(P)$, the middle figure evaluates SAGAN inversion $(SA)$, and the bottom figure evaluates SNGAN inversion $(SN)$.  Each inversion is evaluated on 500 probes created by ProgGAN, SAGAN, and SNGAN.  Note that $S_i \rightarrow 1$ when the algorithm is confident that the $i^{\rm th}$ generator was the source of the probe, and $S_i \rightarrow -1$ when the algorithm is confident generator $i$ was not the source.}
\label{fig:CelebA_histogram}
\end{figure}

\begin{figure}[t]
\begin{center}
\includegraphics[width=0.9\linewidth]{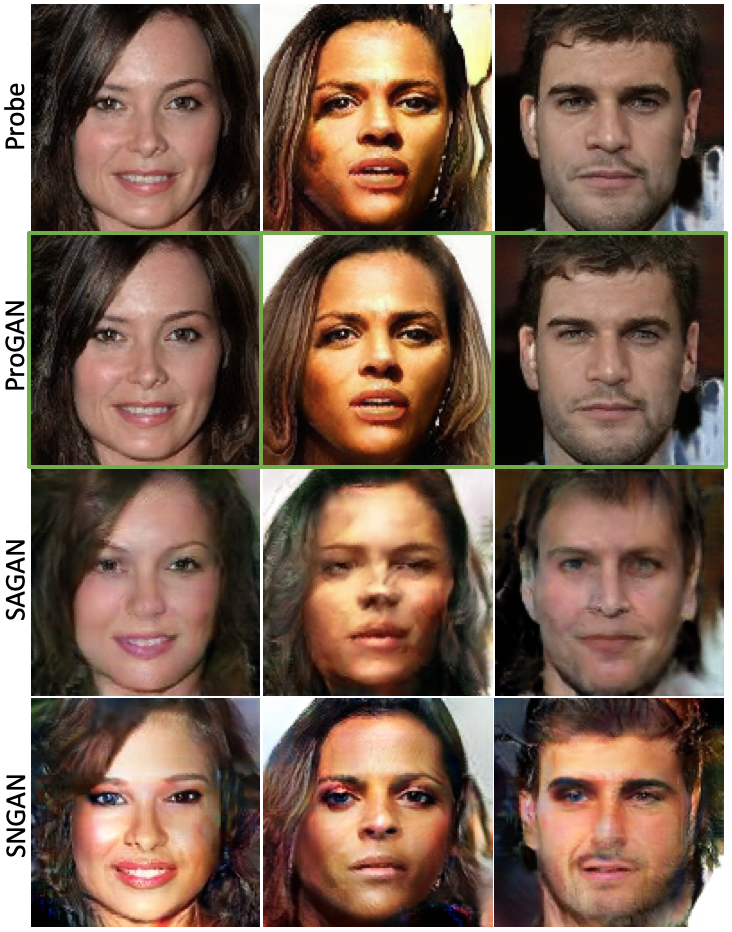} 
\end{center}
\caption{Examples of correct attribution of probe images generated by ProGAN. The top row shows probe images while the $2^{\rm nd}$ through $4^{\rm th}$ rows show the closest approximations found by inverting ProGAN, SAGAN, and SNGAN generators. Green boxes indicate the decision of our attribution algorithm. Note the subtle differences in the 2nd column between the probe and ProGAN reconstruction, indicating an imperfect inversion.}
\label{fig:ProGAN_success}
\end{figure}

\begin{figure}[t]
\begin{center}
\includegraphics[width=0.9\linewidth]{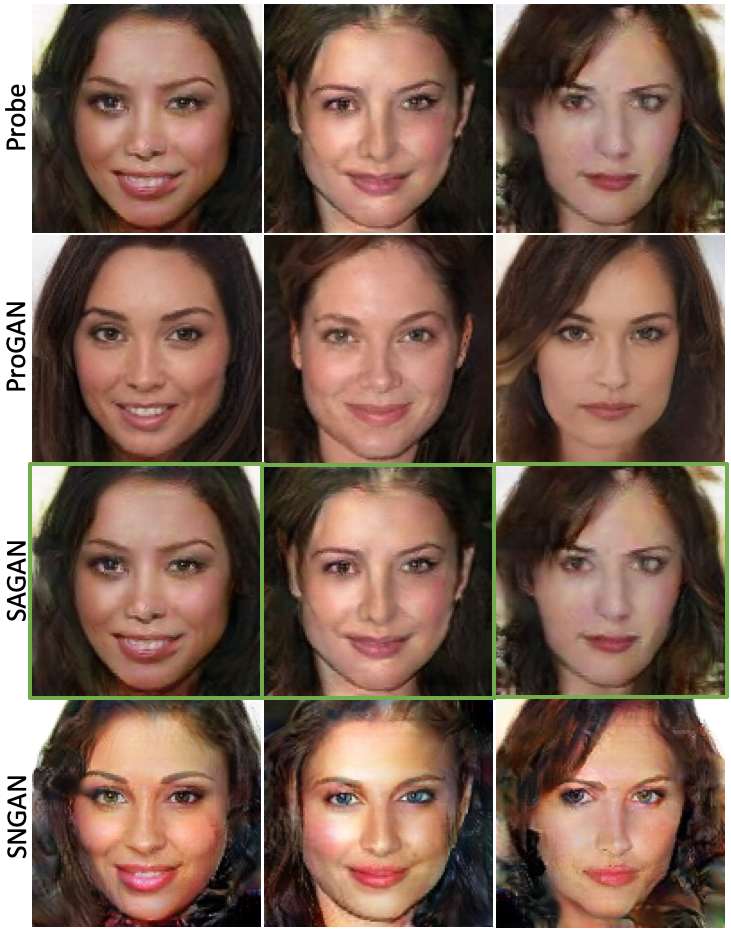} 
\end{center}
\caption{Examples of correct attribution of probe images generated by SAGAN. The top row shows probe images while the $2^{\rm nd}$ through $4^{\rm th}$ rows show the closest approximations found by inverting ProGAN, SAGAN, and SNGAN generators. Green boxes indicate the decision of our attribution algorithm.}
\label{fig:SAGAN_success}
\end{figure}

\begin{figure}[t]
\begin{center}
\includegraphics[width=0.9\linewidth]{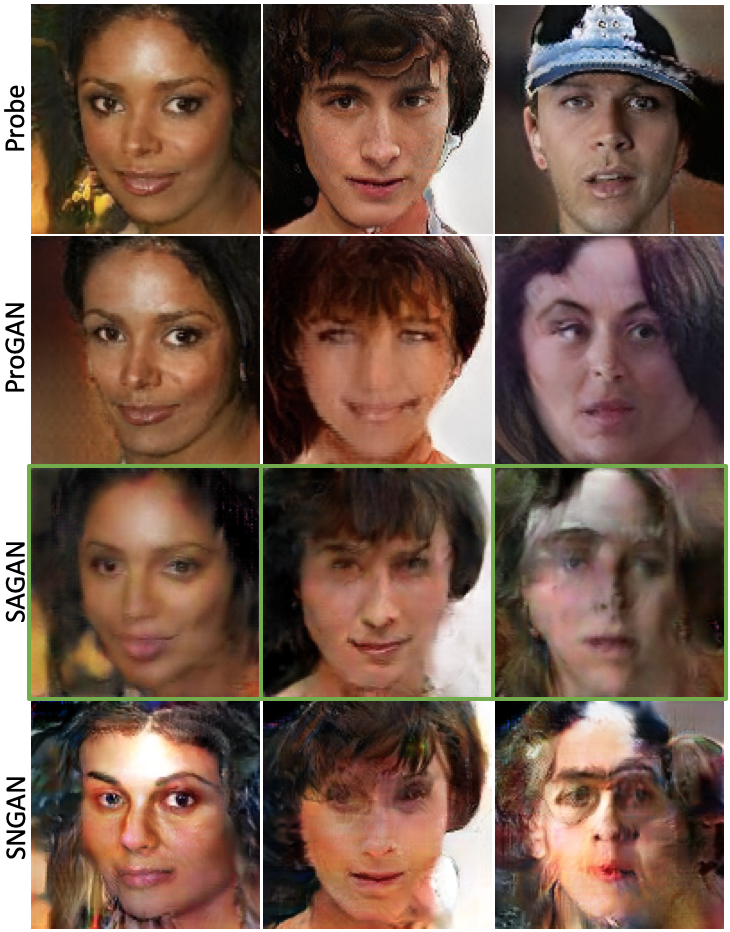} 
\end{center}
\caption{Examples of incorrect attribution of probe images generated by ProGAN. Green boxes indicate the decision of our attribution algorithm.  In each case, mis-attribution is caused by failures of the inversion, which is evident in the ProGAN row in the background in the first column, the face in the second column, and everything in the third column.}
\label{fig:ProGAN_fail}
\end{figure}

We tested inversion-based attribution on a dataset of 1500 synthetic images (500 of which were generated by ProGAN, 500 by SAGAN, and 500 by SNGAN) which were saved as PNGs.  As summarized in Table \ref{table:GANaccuracy}, 
%
attribution accuracy was 100\% for images generated by SAGAN and SNGAN and 95\% for images generated by ProGAN.
%
Examples of successful attributions of images generated by ProGAN and SAGAN are shown in Figures \ref{fig:ProGAN_success} and \ref{fig:SAGAN_success}.  As can be seen in Figures \ref{fig:ProGAN_success} and \ref{fig:SAGAN_success}, successful reconstruction of the probe image by a generator offers compelling, interpretable evidence that a probe image was created by the indicated generator.  This ability to offer strong, interpretable evidence of a synthetic image's source is unique to our method of generator attribution, compared to other attribution methods
which tend to operate as black boxes without strong interpretability or confidence measures.  In Figure \ref{fig:ProGAN_fail}, we show three examples where attribution of images generated by ProGAN failed.  From the figure, it is clear that the cause of the failure is that the optimization algorithm failed to converge to the best reconstruction, causing mis-attribution.  In each of the three cases, it is obvious that the quality of the best image reconstruction is poor, with relatively large residual error.

We can analyze the invertibility of each generator by generalizing Eq. \ref{eq:Sequation} to the situation of more than two generators. We define multiple $S_i$ scores, where each score $S_i$ characterizes the reconstruction error achieved by generator $i$ on a particular image relative to the errors of other generators:
\begin{equation}
 S_i = \frac{  min_{j \ne i}( L^{\rm min}_j ) - L^{\rm min}_i }{  min_{j \ne i}(L^{\rm min}_j) + L^{\rm min}_i}  \, .
 \label{eq:S_i_equation}
\end{equation}
(Intuitively, this is describing one-vs-rest classification.)  Note that with this extended formula, $S_i \rightarrow 1$ when generator $i$ perfectly reconstructs the probe, $S_i \rightarrow -1$ when a different generator $j \ne i$ perfectly reconstructs the probe, and $S_i \rightarrow 0$ when generator $i$'s reconstruction error matches the next-best reconstruction error (i.e., uncertain attribution).  The scores are visualized in Fig. \ref{fig:CelebA_histogram}.  From the bottom two histograms, we can see that inversion-based attribution worked very well for SAGAN and SNGAN---attribution was correct 100\% of the time and with high confidence ($S_i \approx \pm 1 $) on targets from those generators.  From the top histogram, we can see ProGAN attribution was more difficult---there were more errors ($S_P < 0$ on ProGAN targets) and lower attribution confidence ($|S_P| < 1$ on ProGAN targets), indicating that optimization-based inversion of the ProGAN generator on ProGAN targets sometimes struggled to produce extremely low reconstruction errors.

\begin{table}
\begin{center}
\begin{tabular}{ l | c | c | c |}
 \cline{2-4}
 Source    & ProGAN & SAGAN & SNGAN \\ \hline
 Accuracy &   95\%    &  100\% &  100\% \\
 \hline
\end{tabular}
\end{center}
\caption{Inversion-based attribution accuracy, by data source, measured on a dataset of 1500 synthetic images---500 generated by ProGAN, 500 by SAGAN, and 500 by SNGAN.}
\label{table:GANaccuracy}
\end{table}

In addition to the above work, we experimented with minimizing $L1$ loss functions, on the hope that it would encourage a sparser residual error vector $r_i = I - G_i(\tilde{z})$, but we saw no significant attribution benefit. We also tested the LBFGS optimization algorithm,
which uses 2nd-order derivatives to accelerate convergence; similar to \cite{webster2019detecting}, we found that significantly fewer optimization steps were required to invert an image, but we saw no significant benefit in the minimum loss obtained by the optimizer. 

\section{Discussion}

We have shown that generator inversion is a useful means by which to pursue synthetic image attribution, with the added advantage of being able to re-create the generation process by estimating the latent vector via inversion.  Despite this increased functionality, we achieve similar results to black box systems such as Yu \etal \cite{yu_attributing_2018}.  Both our method and Yu's are able to distinguish between generators that are quite similar, up to and including generators trained from the same data in the same order.

\begin{figure}[t]
\begin{center}
\includegraphics[width=0.9\linewidth]{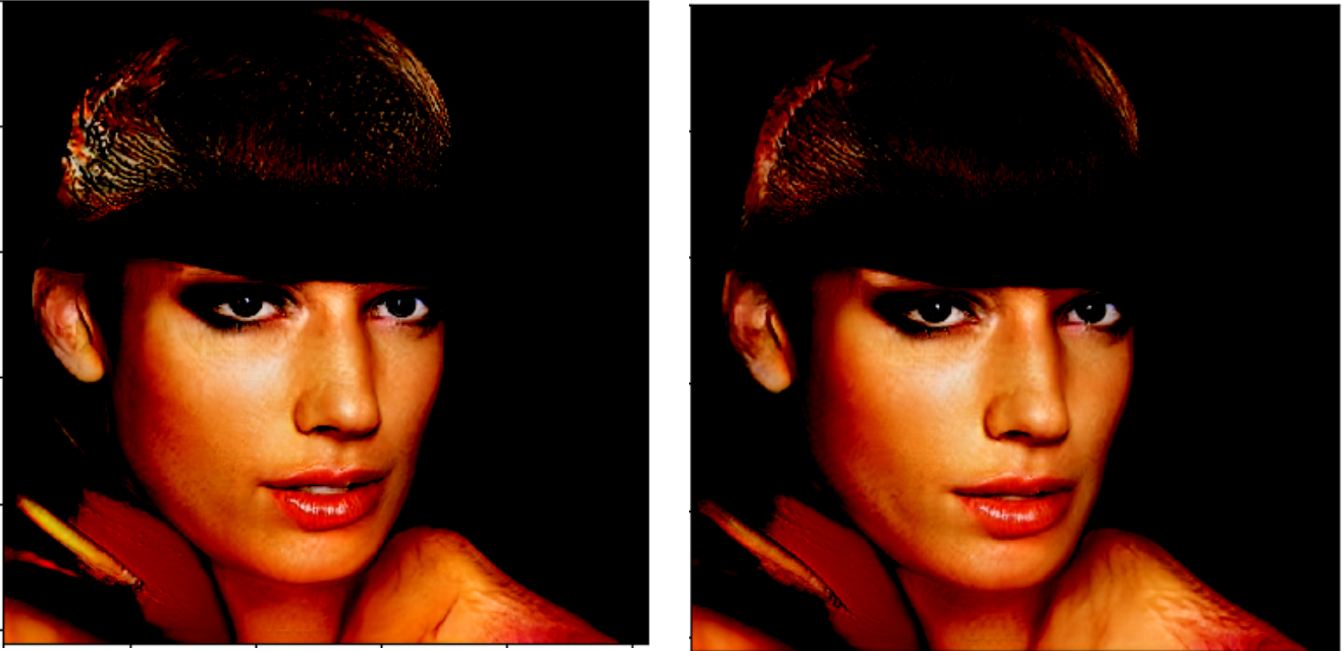} 
\end{center}
\caption{Example inversion with semantically meaningful differences.  In this case, the reconstruction (right) has lips that are slightly more closed and a smaller reflection in the hair (upper left) compared to the probe image (left).  Note that this inversion was performed against the full-resolution (1024x1024) version of the CelebA-HQ ProGAN, whereas inversions in Sec. \ref{subsec:CelebA_experiments} were done against a lower-resolution version.}
\label{fig:semanticDifferences}
\end{figure}

While our attribution results correspond to a completely automated system, it is interesting to consider how our output image would be used and interpreted by a forensic analyst.  As illustrated by Fig. \ref{fig:ProGAN_fail}, inversion failures are visually salient and so the assigned attributions could be rejected by the analyst for failing to meet quality thresholds.  We note, however, that designing such a threshold is non-trivial in light of semantically meaningful differences between the probe and reconstruction in the case of successful inversion, an example of which is illustrated in Fig. \ref{fig:semanticDifferences}.

\section{Future Work}

Though our experiments have so far been limited to attribution between two or three generators, it can easily be extended to $N$-way attribution with the use of Eq. \ref{eq:S_i_equation}; 
note that the dynamic range would need to be normalized for networks having different outputs, i.e. different than the $[-1,1]$ range used by ProGAN, SAGAN, and SNGAN.  However, the computational complexity of inversion over $N$ generators can be quite high.  Given the computational complexity of inverting the generator, a more efficient system could use an attribution-only method, i.e. one of \cite{yu_attributing_2018,marra_attributing_2018}, to determine the source generator and then apply inversion on only that generator.  Another extension would be to train neural network encoders to explicitly perform the inverse mapping from an image to a latent vector, which could also be further fine-tuned by optimization, as in \cite{zhu_generative_2016}.

One area of future work would be to extend our attribution capabilities into the black box domain, in order to handle generators which may be offered online as a service, but without published network weights.  It has been shown that, for deep networks performing classification tasks, representative proxy networks can be trained based on a relatively small number of input/output pairs from the target network.  It may be the case that we can develop proxies for target {\em generators} in order to support our white-box attribution method.  

\ifcvprfinal
\section*{Thanks}
We thank Ning Yu for sharing data and pre-trained face generators used in this work, and Asongu Tambo for helpful suggestions.  

\section*{Acknowledgements}
This research was developed with funding from the Defense Advanced Research Projects Agency (DARPA). The views, opinions and/or findings expressed are those of the author and should not be interpreted as representing the official views or policies of the Department of Defense or the U.S. Government.
\fi

{\small
\bibliographystyle{ieee_fullname}
\bibliography{GAN-Inversion_zoteroexport}
}
\end{document}